%% file: root.tex
\documentclass[letterpaper, 10 pt, conference]{ieeeconf}  

\IEEEoverridecommandlockouts                              

\usepackage{graphicx} 
\usepackage{stfloats} 
\graphicspath{{figures/}}
\usepackage{epsfig} 
\usepackage{mathptmx} 
\usepackage{times} 
\usepackage{amsmath} 
\usepackage{amssymb}  
\def\BibTeX{{\rm B\kern-.05em{\sc i\kern-.025em b}\kern-.08em
    T\kern-.1667em\lower.7ex\hbox{E}\kern-.125emX}}
\usepackage{enumerate}
\usepackage{subcaption}
\usepackage{cite}
\usepackage[table,xcdraw]{xcolor}

\definecolor{PineGreen}{RGB}{0,153.0,0}
\definecolor{LightCyan}{rgb}{0.9,0.9,1}

\usepackage{array}

\usepackage{pifont}
\newcommand{\cmark}{\ding{51}}%
\newcommand{\xmark}{\ding{55}}%

\DeclareMathOperator{\clip}{clip}

\newcommand{\sev}{\alpha}                 
\newcommand{\sevhat}{\hat{\alpha}}
\newcommand{\gd}{g_{d}}                   
\newcommand{\pih}{\pi^{h}}                
\newcommand{\pid}{\pi^{d}}                
\newcommand{\dres}{\Delta^{d}}            
\newcommand{\act}{u}                      
\newcommand{\aconv}{a}                    
\newcommand{\phisec}{\phi}                
\newcommand{\phioff}{\phi^{\mathrm{off}}} 
\newcommand{\phibias}{\Phi}               
\newcommand{\sgcpg}{\mbox{SG-CPG}}     
\newcommand{\Zvec}{\mathbf{Z}}            
\newcommand{\Zhat}{\hat{\mathbf{Z}}}      
\newcommand{\SI}[2]{\ensuremath{#1\,\mathrm{#2}}}  
\newcommand{\todo}[1]{}  

\title{\LARGE \bf
SG-CPG: \underline{S}everity-\underline{G}ated \underline{C}entral \underline{P}attern \underline{G}enerators for Adaptive Quadruped Locomotion under Continuous Actuator Degradation
}

\author{Adarsh Kumar Kosta and Kaushik Roy
\thanks{Adarsh Kumar Kosta is a PhD Candidate at the Department of Electrical and Computer Engineering, Purdue University, West Lafayette, IN 47906, USA
        {\tt\small akosta@purdue.edu}}%
\thanks{Kaushik Roy is Edward G. Tiedemann Jr. Distinguished Professor at the Department of Electrical and Computer Engineering, Purdue University, West Lafayette, IN 47906, USA
        {\tt\small kaushik@purdue.edu}}%
}

\IEEEaftertitletext{%
\vspace{-0.6\baselineskip}%
\begin{minipage}{\textwidth}%
\centering%
\includegraphics[width=0.98\textwidth]{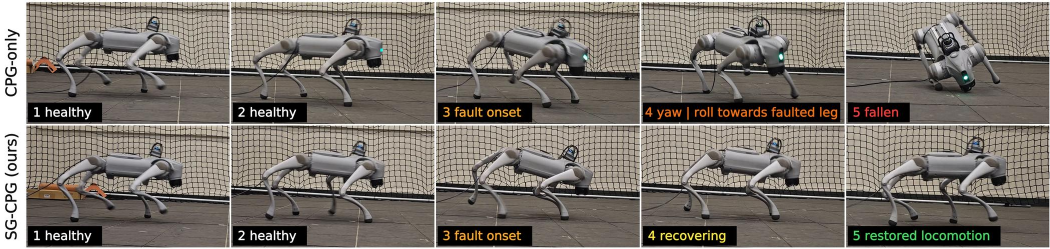}%
\captionof{figure}{A Unitree Go2 moving at \SI{0.6}{m/s}. At fault onset, the front-right (FR) calf's torque ceiling drops by  $92\%$, $\sev = 0.92$.
(Row1) CPG-only yaws/rolls towards the faulted leg and falls.
(Row2) \sgcpg{} engages its residual to restore locomotion.}%
\label{fig:teaser}%
\vspace{1mm}
\end{minipage}%
}

\begin{document}

\maketitle
\thispagestyle{empty}

\pagestyle{empty}

\begin{abstract}
An animal with a weakened limb does not necessarily switch its gait, instead it unloads the affected limb, re-coordinates the remaining limbs, and scales its response with injury severity. This graded adaptation allows locomotion to persist despite partial loss of limb strength, rather than requiring a discrete transition between healthy and failed. Inspired by this behavior, we propose SG-CPG, a central pattern generator (CPG) for quadruped locomotion under continuous actuator degradation. SG-CPG preserves a frozen healthy CPG policy and introduces two severity-driven gates: a residual gate that re-coordinates all four legs and an amplitude gate that progressively shortens the weakened leg's stride as degradation increases. We emulate progressive degradation through two mechanisms: lowering the joint torque ceiling (\texttt{ceiling}) and scaling its low-level controller gains (\texttt{gain}), representing distinct forms of actuator weakening. Our simulations on a Unitree Go2 show that SG-CPG maintains a trot gait with $100\%$ survival across an omnidirectional command schedule under $95\%$ joint strength loss while tracking commands within $8\%$. Under a lowered torque ceiling, removing either severity path, the residual's severity observation or the amplitude gate, raises clipping at the weakened joint from $4.4\%$ to $13.6\%$ and $26.3\%$ of steps at an $80\%$ loss. On a real Go2, SG-CPG survives $28$ of $29$ forward and turning trials with up to $93\%$ calf torque degradation. These results show that severity-gated adaptation can extend a healthy locomotion policy to progressive actuator degradation without treating the fault as a discrete failure.

\end{abstract}

\input{sections/01_introduction}

\input{sections/02_background_and_related_work}

\input{sections/03_methodology}
\input{sections/04_experimental_setup}
\input{sections/05_experiments_and_results}

\input{sections/06_hardware}

\input{sections/07_conclusion_and_futurework}
\bibliography{root.bib}
\bibliographystyle{ieeetr}   

\end{document}

%% file: sections/01_introduction.tex
\section{Introduction} \label{sec:intro}

Rhythmic locomotion in vertebrates is generated in large part by spinal circuits rather than being explicitly orchestrated by the brain~\cite{brown1911intrinsic, kiehn2016decoding}. Central pattern generators (CPGs), which underlie this rhythmic activity, can sustain locomotor oscillations without rhythmic descending input, while descending commands and sensory feedback regulate the resulting motion~\cite{grillner2020current, rossignol2006dynamic}. This organisation provides a compact control structure in which a small number of signals regulate locomotion while the underlying dynamics maintain inter-leg coordination. Importantly, this coordination adapts to physical impairment. Cats with denervated ankle muscles recover stance and locomotor function by adapting their existing locomotor pattern~\cite{carrier1997neurectomy, pearson1999denervation, bouyer2001plasticity}. Previous work~\cite{chang2009wholelimb} shows that, after peripheral nerve injury, whole-limb function can be preserved through coordinated changes across joints, with recovery developing progressively over time~\cite{chang2018progressive}. Dogs with lameness reduce load on the affected limb while redistributing load across the remaining limbs~\cite{goldner2018lameness}, while insects and spiders re-coordinate their remaining legs after limb loss~\cite{hughes1957coordination, kane2025spiders}. A similar pattern appears with gradual age-related decline, with human gait changing as strength, balance, and stability degrade~\cite{cruz2017gait}. Together, these observations show that locomotor impairment can range from discrete events, such as injury or limb loss, to continuously graded changes, such as ageing, with adaptation occurring through rapid re-coordination or progressive adjustment. This suggests that locomotion can accommodate continuously varying impairment through correspondingly graded adaptation.

Robotic CPGs provide a natural framework for capturing key properties of biological CPGs. They generate rhythmic trajectories around stable limit cycles while exposing interpretable parameters for regulating locomotion~\cite{ijspeert2008review, ijspeert2014biorobotics}. Recent CPG-based quadruped controllers combine these dynamics with deep reinforcement learning to learn inter-leg coordination~\cite{bellegarda2022cpgrl, bellegarda2024allgaits, bellegarda2024visual, shafiee2024viability}. In foot-position CPGs, oscillator amplitude directly controls the spatial extent of a leg's motion~\cite{bellegarda2022cpgrl}, providing an interpretable interface for progressive actuator degradation: the weakened leg can operate with reduced motion amplitude while the remaining limbs re-coordinate~\cite{bellegarda2024allgaits}.

This capability is particularly relevant during long-duration deployment, where actuator strength can change with temperature, wear, and sustained loading~\cite{bellicoso2018realworld, gehring2021anymalhvdc}. Thermal effects can reduce available joint torque~\cite{wan2026thermal, qian2026thermalaware}, while motor degradation can reduce the available torque-speed envelope~\cite{wensing2017proprioceptive, shin2025motorregion}. Most fault-tolerant locomotion methods instead consider discrete failures such as locked or masked joints~\cite{liu2022limping, kim2024masking}. Continuous degradation has also been modelled through gain degradation and actuator saturation~\cite{farid2018fractional}, but progressive strength loss presents a distinct problem: the actuator remains functional while its maximum available torque changes continuously.

We study this problem using two degradation mechanisms: capping the torque ceiling (\texttt{ceiling}) and applying low-level gain scaling (\texttt{gain}). These mechanisms expose an important observability distinction. With \texttt{ceiling}, commands below the reduced limit are delivered normally, making a weak joint indistinguishable from a healthy, lightly loaded joint until the ceiling binds. Degradation therefore becomes observable only when the gait sufficiently excites the actuator. With \texttt{gain}, controller output is attenuated at each step, removing this dead band. This distinction matters because reducing the weakened leg's stride lowers its torque demand but may also reduce the excitation available for estimating its remaining strength.

These observations impose two requirements for progressive fault adaptation. The response should scale continuously with remaining actuator strength rather than switch between discrete fault modes, and adaptation should be distributed across the locomotor system because reducing one leg's contribution changes coordination and load sharing in the others. Existing approaches address these requirements separately. Fault-tolerant CPG controllers handle declared joint faults or leg failures~\cite{zhang2025ftcpg, ren2015chaotic}, and CPG policies can walk with one or two legs disabled without retraining~\cite{bellegarda2024allgaits}. Other approaches model continuous actuator weakening in joint space~\cite{luo2023ftnet, wu2023adapt, lee2025dreamflex, gravina2026gaittiming}, while implicit adaptation methods infer dynamics from proprioceptive histories~\cite{kumar2021rma, lee2020challenging}. For example, RMA trains over $10\%$ motor-strength variation rather than explicitly representing continuously varying remaining torque capacity. Under a torque-ceiling fault, moreover, actuator-strength changes may be absent from observations until the actuator binds. These approaches therefore do not directly address the combination of continuous severity-dependent response, distributed CPG re-coordination, and actuator-strength observability under a torque ceiling.

We address these requirements with \sgcpg{}, a severity-gated CPG controller for quadruped locomotion with a single weakened joint over flat ground. Starting from a frozen healthy CPG policy $\pih$, SG-CPG introduces two severity-dependent mechanisms. A residual gate activates a single learned residual $\pid$ when a joint weakens and applies it across all four legs, allowing the remaining limbs to re-coordinate. An amplitude gate simultaneously reduces the affected leg's stride amplitude as severity increases, reducing its torque demand. At zero \emph{reported} severity, both gates are inactive and the composed controller is exactly $\pih$. The controller therefore separates whole-body re-coordination from reduction of the weakened limb's demand while retaining an interpretable CPG representation.

\input{tables/t1_related_work}

We compare SG-CPG against the frozen $\pih$, a severity-blind residual, and a variant without the amplitude gate. We also evaluate analytic and learned severity estimators under torque-ceiling and gain-scaling faults to characterise observability. On hardware, severity is hand-set and the torque ceiling is approximated by clamping position error. Our contributions are:

\begingroup
\setlength{\leftmargini}{13.3pt}

\begin{enumerate}
\item We introduce \sgcpg{}, a severity-gated CPG controller that extends a frozen healthy locomotion policy to continuous actuator degradation through severity-dependent whole-body coordination and faulted-leg stride reduction.


\item We consider two severity-graded actuator fault mechanisms, \texttt{ceiling} and \texttt{gain}, and develop fault-residual policies for locomotion recovery under each fault.


\item We demonstrate the importance of the residual and severity gates through ablations and characterise the observability of both degradation mechanisms, including the dead band induced by torque-ceiling faults.


\item We validate \sgcpg{} in simulation and on a Unitree Go2, demonstrating transfer to hardware with $28$ of $29$ forward and turning trials surviving under up to $93\%$ emulated calf torque degradation. The amplitude gate alone cuts clamped calf ticks from $77\%$ to $39\%$ and mean body tilt from $7.0^\circ$ to $2.8^\circ$.

\end{enumerate}
\endgroup

%% file: tables/t1_related_work.tex
\begin{table}[t]
\centering
\scriptsize
\setlength{\tabcolsep}{3pt}
\renewcommand{\arraystretch}{1.15}
\caption{\sgcpg{} against CPG and fault-tolerant works.}
\label{tab:related}
\begin{tabular}{@{}lccll@{}}
\hline \rowcolor{gray!20}
\textbf{Method} & \textbf{CPG} & \textbf{Graded} & \textbf{Fault model} & \textbf{Fault input} \\
\hline
Ren et al.~\cite{ren2015chaotic} & \cmark & \xmark & Disabled legs & None \\
FT-CPG~\cite{zhang2025ftcpg} & \cmark & \xmark & Locked, power loss & Joint class \\
FT-Net~\cite{luo2023ftnet} & \xmark & \cmark & Gain scaling & Implicit latent \\
DreamFLEX~\cite{lee2025dreamflex} & \xmark & \cmark & Locked, torque scale & Binary, latent \\
\hline
\rowcolor{blue!10}
\textbf{\sgcpg{} (ours)} & \cmark & \cmark & Ceiling, gain scaling & Severity $\sev_{ij}$ \\
\hline
\end{tabular}
\vspace{-4mm}
\end{table}

%% file: sections/02_background_and_related_work.tex
\section{Related Work} \label{sec:brw}

\subsection{CPG-based locomotion}

Central pattern generators provide an interpretable representation of rhythmic locomotion through oscillator phase, amplitude, frequency, and inter-leg coupling. CPG-RL learns these parameters jointly with a policy; each leg uses an amplitude-controlled oscillator, with the policy coordinating the legs rather than learned inter-oscillator coupling~\cite{bellegarda2022cpgrl}. AllGaits instead uses a user-selected coupling matrix while learning leg-specific amplitude and frequency modulation~\cite{bellegarda2024allgaits}. Neither approach considers actuator degradation. \sgcpg{} retains the coupling structure of a frozen healthy policy $\pih$ and introduces degradation-dependent changes through a learned residual and oscillator amplitude, allowing severity to modify both whole-body coordination and the affected leg's motion demand.

\begin{figure*}[t]
\centering
\includegraphics[width=0.95\textwidth]{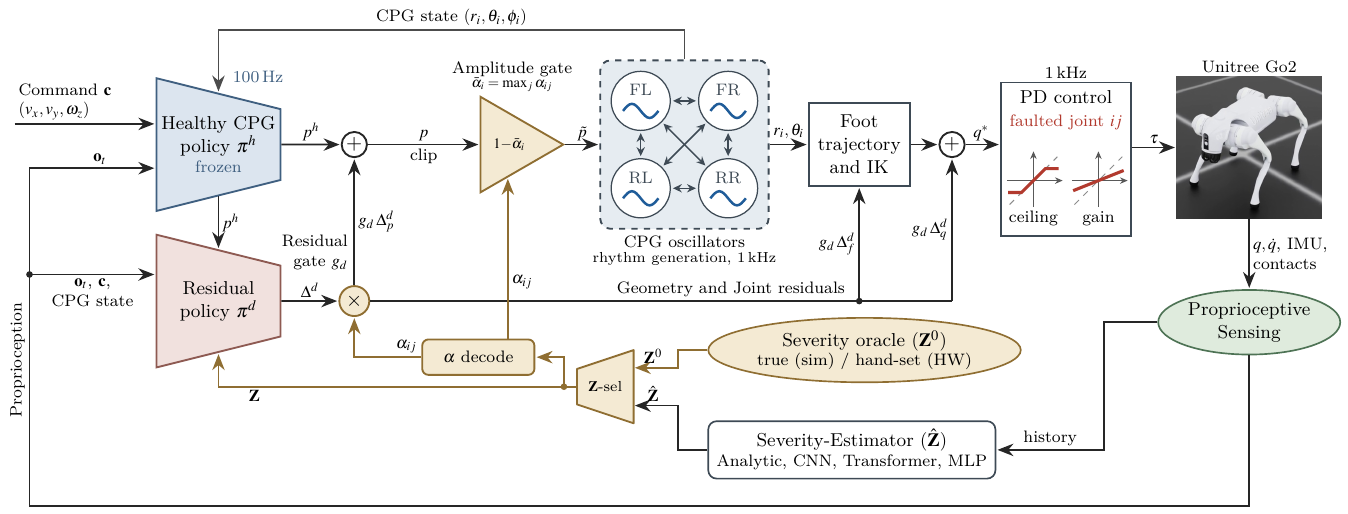}
\caption{\sgcpg{} control architecture: a frozen CPG policy $\pih$ emits $p^h$. A residual $\pid$ corrects $p^h$, the foot geometry $\dres_{f}$, and the joint targets $\dres_{q}$ when the gate $\gd$ opens~\eqref{eq:compose}.
The $\sev$ decode block feeds both gates, and $\Zvec$-sel passes $\Zvec^{0}$ or $\Zhat$.}
\label{fig:arch}
\vspace{-2mm}
\end{figure*}

\subsection{Fault-tolerant locomotion}

Fault-tolerant locomotion has been addressed through damage-specific search, self-modeling, behavioral selection, and adaptive CPG dynamics~\cite{bongard2006resilient, cully2015adapt, koos2013tresilience, ren2015chaotic}. More recent learning-based approaches often treat failure as a discrete event through locked or masked joints, history-based fault inference, or fault-specific behaviors~\cite{liu2022limping, kim2024masking, hou2024multitask, xu2025acl}. FT-CPG augments a frozen CPG with fault-residual policies, training a separate residual for each affected leg~\cite{zhang2025ftcpg}.

Other methods model actuator weakening continuously. FT-Net, DreamFLEX, and adaptive gait-timing approaches introduce degradation during training, with evaluations focused on substantial weakening or recovery near joint failure~\cite{luo2023ftnet, lee2025dreamflex, gravina2026gaittiming}. ADAPT provides a continuous degradation level as prior information, but models gain scaling rather than a torque ceiling and does not use a CPG representation through which severity shapes the gait~\cite{wu2023adapt}. \sgcpg{} instead uses a single residual shared across all four legs, allowing a weakened joint to trigger whole-body re-coordination rather than selection of a leg-specific policy. Its amplitude gate separately reduces the affected leg's stride and torque demand as severity increases. Unlike FT-CPG, which detects concurrent faults within one leg online, \sgcpg{} is evaluated with the severity of a single weakened joint provided as true or hand-set information.

\subsection{Implicit adaptation and residual policies}

Implicit adaptation methods infer changes in terrain or robot dynamics from proprioceptive histories~\cite{lee2020challenging, kumar2021rma}. RMA includes wear and tear among the conditions to which it adapts, while randomizing motor strength over a roughly $10\%$ range during training~\cite{kumar2021rma}. Under a torque-ceiling fault, however, reduced capacity has no observable effect until the commanded torque reaches the ceiling (Sec.~\ref{sec:results:est}).

Residual policy methods add a learned correction to a fixed controller and initialize it at zero, but do not constrain it to remain zero after training~\cite{silver2018residual}. Wan et al.~freeze a nominal policy while softly preserving its behavior through an $\ell_2$ penalty on the residual~\cite{wan2026thermal}. \sgcpg{} instead structurally preserves $\pih$ at zero severity: the residual gate is zero and the amplitude gate leaves $\mu$ unchanged, with adaptation progressively admitted as severity increases.

%% file: sections/03_methodology.tex
\section{Methodology} \label{sec:method}

A controller for a weakening robot must keep the healthy gait intact, correct it once a joint weakens, and ask less of the weakened leg.
\sgcpg{} assigns these tasks to a frozen healthy CPG policy $\pih$, a residual policy $\pid$ admitted by a residual gate $\gd$, and an amplitude gate (Fig.~\ref{fig:arch}).

\subsection{The CPG formulation}
\label{sec:method:cpg}

\begin{figure}[t]
\centering
\includegraphics[width=\columnwidth]{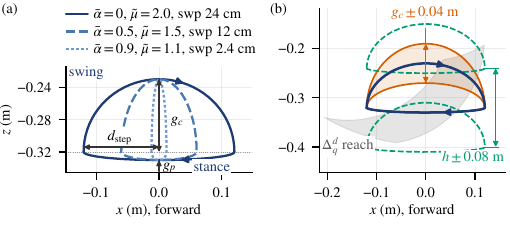}
\caption{Commanded foot targets from Eq.~\eqref{eq:foot}. (a)~stretched in $z$, (b)~to scale.
The gate shortens the sweep at fixed clearance. The residual shifts $g_c$, $h$, and the joints within grey.}
\label{fig:footmap}
\vspace{-4mm}
\end{figure}

Each leg $i$ carries one amplitude-controlled phase oscillator~\cite{bellegarda2022cpgrl} with amplitude $r_i$, phase $\theta_i$ and secondary phase $\phisec_i$, driven by a commanded amplitude $\mu_i$, frequency $\omega_i$ and secondary-phase rate $\psi_i$ (in Hz),
\begin{equation}
\begin{aligned}
\ddot r_i &= \aconv\bigl(\tfrac{\aconv}{4}(\mu_i - r_i) - \dot r_i\bigr), \qquad \dot\phisec_i = 2\pi\psi_i, \\
\dot\theta_i &= 2\pi\omega_i + \textstyle\sum_{k} r_k\, w_k \sin\bigl(\theta_k - \theta_i - \phibias_{ik}\bigr),
\end{aligned}
\label{eq:osc}
\end{equation}
with convergence constant $\aconv = 150$ as in CPG-RL and phase bias $\phibias_{ik} = \phioff_k - \phioff_i$.
The policy also emits the phase offsets $\phioff_i$ that choose the gait and a coupling weight $w$, shared across legs in $\pih$.
Oscillator state maps to a foot target in the hip frame,
\begin{equation}
\begin{aligned}
x_i &= -d_{\mathrm{step}}(r_i - 1)\cos\theta_i\cos\phisec_i, \\
y_i &= y^0_i - d_{\mathrm{step}}(r_i - 1)\cos\theta_i\sin\phisec_i, \\
z_i &= -h + \begin{cases} g_c \sin\theta_i, & \sin\theta_i > 0 \ \text{(swing)}, \\ g_p \sin\theta_i, & \text{otherwise (stance)}, \end{cases}
\end{aligned}
\label{eq:foot}
\end{equation}
with step length $d_{\mathrm{step}}$ and lateral stance offset $y^0_i$, while stance depth $h$, swing clearance $g_c$ and stance penetration $g_p$ shape the vertical path (Fig.~\ref{fig:footmap}a).
The foot sweeps with amplitude $d_{\mathrm{step}}(r_i - 1)$, so $\mu_i$ is a reach command and a natural place for severity to act.
The policy's raw output $\act^h \in \mathbb{R}^{17}$ decodes to the CPG parameters $p^h$, which stack $[\mu_i, \omega_i, \psi_i]$ for each leg with the four phase offsets and the coupling weight $w$ (Sec.~\ref{sec:setup}).

\subsection{Severity: two mechanisms that are not interchangeable}
\label{sec:method:severity}

Severity $\sev_{ij} \in [0, 1)$ grades joint $j$ of leg $i$, zero when healthy and capped at $0.99$ throughout.
A healthy joint under PD control delivers
\begin{equation}
\tau_{ij} = \clip\bigl(\tau^{\mathrm{PD}}_{ij},\; \pm L_j\bigr), \qquad \tau^{\mathrm{PD}}_{ij} = K_{p,j}\,(q^{*}_{ij} - q_{ij}) - K_{d,j}\,\dot q_{ij} ,
\label{eq:pd}
\end{equation}
where $q_{ij}$ is the joint position, $q^{*}_{ij}$ its target, and the gains and nominal limit $L_j$ are fixed by the joint type.
An actuator can weaken at either of two places in this law, and the place decides what proprioception sees.

A \textbf{torque ceiling} (\texttt{ceiling}) keeps the PD gains and lowers the torque limit,
\begin{equation}
\tau_{ij} = \clip\bigl(\tau^{\mathrm{PD}}_{ij},\; \pm(1-\sev_{ij})\, L_j\bigr),
\label{eq:ceiling}
\end{equation}
and in simulation it scales the whole DC-motor torque--speed envelope that peaks at $L_j$.
While the PD demand stays within the lowered limit the joint moves exactly as a healthy one would, so the fault sits in a dead band that proprioception cannot see, which is why \texttt{ceiling} carries our main results.

\textbf{Gain scaling} (\texttt{gain}) instead scales the PD torque,
\begin{equation}
\tau_{ij} = \clip\bigl((1-\sev_{ij})\,\tau^{\mathrm{PD}}_{ij},\; \pm L_j\bigr),
\label{eq:gain}
\end{equation}
so the torque changes on every step and no severity leaves a dead band.
This is FT-Net's effectiveness model~\cite[Eq.~(7)]{luo2023ftnet}, whose coefficient equals $1 - \sev_{ij}$.

\subsection{Gated residual composition}
\label{sec:method:residual}

We freeze $\pih$, so learning a fault response cannot erode the healthy gait.
$\pid$ corrects each per-leg CPG parameter $p \in \{\mu, \omega, \psi, \phioff, w\}$, the foot geometry and each joint target through one shared gate,
\begin{equation}
\begin{aligned}
p_i &= \clip\bigl(p_i^h + \gd\,\dres_{p,i}\bigr), \qquad \gd = \mathbf{1}\bigl[\max\nolimits_{i',j'} \sev_{i'j'} > 0\bigr], \\
g_i &= \clip\bigl(g^{0} + \gd\,\dres_{f,i}\bigr), \qquad g \in \{g_c, h, \beta\}, \\
q^{*}_{ij} &= \mathrm{IK}_j(x_i, y_i, z_i) + \gd\,\dres_{q,ij} ,
\end{aligned}
\label{eq:compose}
\end{equation}
where $p_i^h$ is the matching entry of $p^h$, each foot-geometry channel $g$ starts from the fixed healthy value $g^{0}$ of Table~\ref{tab:setup}, $\dres$ denotes $\pid$'s output, and the stance fraction $\beta$ re-times swing and stance without moving the foot path.
Each clip keeps $\mu$, $\omega$ and $\psi$ inside the ranges $\pih$ already commands, while $\phioff$ wraps and the coupling weight's floor falls to $0$, letting a weak leg decouple.
The geometry clips never bind, since $g_c$, $h$ and $\beta$ reach only Table~\ref{tab:setup}'s $0.09 \pm 0.04$~m, $0.32 \pm 0.08$~m and $0.5 \pm 0.25$.
The residual gate reads the severities $\sev_{ij}$ alone and emits one bit, which admits or blocks every channel of $\dres$ at once.
That gated residual enters at three places (Fig.~\ref{fig:arch}): twenty channels of $\dres_{p}$ join $p^h$ before the oscillators, twelve of $\dres_{f}$ reshape the foot path, and twelve of $\dres_{q}$ offset the joint targets after inverse kinematics.
The gate is deliberately global, because a per-leg mask would zero the residual on the healthy legs that must compensate.

$\pih$ observes $68$ proprioceptive, command, CPG-state and contact inputs with its own previous action.
$\pid$ observes those same $68$ with its own previous $44$ outputs, $p^h$, and the fault vector $\Zvec \in \mathbb{R}^{24}$, twelve lock flags that stay zero in this work followed by the twelve severities $\sev_{ij}$.
It is trained with PPO on the simulator's true (\emph{oracle}) severity in four arms that cross both mechanisms with the two healthy bases of Sec.~\ref{sec:setup}: the default observation space, giving \sgcpg{}, and ZBLV, which zeroes the base linear velocity, giving \sgcpg{}$_{\mathrm{ZBLV}}$.

\subsection{The severity-to-amplitude gate}
\label{sec:method:gate}

The amplitude gate asks less of a weak leg by acting on the oscillator directly, after the composed $\mu_i$ is clipped,
\begin{equation}
\tilde\mu_i = \mu_{\min} + (1 - \bar\sev_i)\,(\mu_i - \mu_{\min}), \qquad \bar\sev_i = \max_{j} \sev_{ij} ,
\label{eq:gate}
\end{equation}
and $\tilde\mu_i$ replaces $\mu_i$ in~\eqref{eq:osc}, so the gated parameters $\tilde p$ that reach the oscillators differ from $p$ in $\mu_i$ alone.
With $\mu_{\min} = 1$ the gate scales the steady-state sweep of~\eqref{eq:foot} by exactly $1 - \bar\sev_i$ and leaves the swing clearance unchanged (Fig.~\ref{fig:footmap}a).
A shorter sweep lowers the torque this joint must produce, so a lowered cap binds on fewer of its steps.
Nothing prescribes how this relief should grow, so we use the simplest monotone form and let the residual correct around it.

\subsection{Estimating severity}
\label{sec:method:est}

Because the policies read the true severity in training, we ask which estimator could supply it.
The \textbf{analytic} estimator reads severity off the peak torque in a window, $\sevhat_{ij} = 1 - \max_t |\tau_{ij}| / L_j$.

The \textbf{learned} estimators re-implement the FT-Net and FT-CPG architectures and add two MLP controls, all with \emph{identical} inputs and head, so any difference lies in the temporal stage.
Each \sgcpg{} arm is rolled out under training commands, faulting $90\%$ of episodes, and each dataset is split by \emph{episode} with $20\%$ held out.
Every step contributes the $68$ inputs of $\pih$ other than its previous action, read as a suffix of one shared $30$-step window.
\textbf{CNN-30}~\cite{luo2023ftnet} convolves over $30$ steps and \textbf{\mbox{Transformer-10}}~\cite{zhang2025ftcpg} attends over the last ten, since saturation is intermittent and scattered through the window.
\textbf{MLP-30} flattens the same $30$ steps with more parameters than CNN-30, so the comparison measures architecture rather than capacity, and \textbf{MLP-1} reads only the latest step.
A shared head reads out $12$ lock logits (BCE) with $12$ severity means and log-variances (\mbox{$\beta$-NLL}).
Predicting $\Zvec$ rather than a latent lets each model's output $\Zhat$ stand in for the oracle's vector without changing the controller.

%% file: sections/04_experimental_setup.tex
\section{Experimental Setup} \label{sec:setup}

\begin{table}[t]\centering\footnotesize
\renewcommand{\arraystretch}{1.15}
\caption{Policy settings and parameter ranges.}
\label{tab:setup}
\setlength{\tabcolsep}{5pt}
\begin{tabular}{lcc}\hline
 & $\pih$ & $\pid$ \\ \hline
observation, action (dim.) & $85$, $17$ & $153$, $44$ \\ \hline
$\mu_i$ & $[1, 2.5]$ & $\pm 1$ \\
$\omega_i$ (Hz) & $[0, 4]$ & $\pm 1$ \\
$\psi_i$ (Hz) & $\pm 1.5$ & $\pm 3$ \\
coupling weight $w$ & $[1, 2]$ & $\pm 2$ \\ \hline
$g_{c,i}$, $h_i$ (m); $\beta_i$ & $0.09$, $0.32$; $0.5$ & $\pm 0.04$, $\pm 0.08$; $\pm 0.25$ \\
$\dres_{q, hip}$, $\dres_{q, thigh}$, $\dres_{q, calf}$ (rad) & -- & $\pm 0.15/0.19/0.68$ \\ \hline
friction & $[0.5, 1]$ & $[0.5, 1]$ \\
mass (kg) & $[0, 5]$ & $[0, 5]$ \\
push (m/s) & $\pm 0.5$ & $\pm 0.5$\\ \hline
\end{tabular}
\vspace{-4mm}
\end{table}

\subsection{Simulation and training}
We train in Isaac Lab on a Unitree Go2 over flat ground, in \SI{20}{s} episodes ending when the base, head or a thigh touches the ground.
Both policies act at \SI{100}{Hz} over a \SI{1}{kHz} physics step with \SI{0.03}{kg\,m^2} armature, using CPG-RL's networks and PPO hyperparameters~\cite{bellegarda2022cpgrl}.
PD gains $K_p$ are $119/93/50$ and $K_d$ are $2.26/1.76/0.62$ against nominal torque caps $L_j$ of $23.70/23.70/\SI{45.43}{N\,m}$, and training commands are drawn every \SI{10}{s} from $v^*_x \in [-1, 1.5]$, $v^*_y \in [-1, 1]$ and $\omega^*_z \in [-1, 1]$.
Rewards resemble CPG-RL's, combining velocity tracking with penalties on vertical and roll--pitch base motion, joint work, raw action excess and foot slip. $\pid$ drops the height and tilt penalties used by $\pih$ and alone receives a stride-amplitude bonus. Only $\pih$ sees PD gain scaling, $\mathcal{U}(0.8, 1.2)$. The second healthy base, \textbf{ZBLV}, differs only by zeroing the base-linear-velocity observation of both policies during training. Its residuals also see the friction, mass, and push ranges of Table~\ref{tab:setup}, whereas the \sgcpg{} residuals train without randomisation.
Each training episode weakens one uniformly drawn joint to a constant severity at a moment within the first \SI{3}{s}. Severity is drawn uniformly from each joint's clip onset to $0.99$, since including zero would make a ceiling fault nearly inert in $35.4\%$ of \sgcpg{} episodes.

\subsection{Evaluation protocol}
\label{sec:protocol}
We pair survival with velocity tracking wherever possible, since stopping and balancing alone does not demonstrate locomotion. Each arm is one training seed, run with its mean action in three repeats per cell from random phases. Each cell weakens one joint to a fixed severity from \SI{4}{s}; any fall before the episode ends counts against survival.

The \textbf{straight-line sweep} runs for \SI{20}{s} across the $12$ joints and $13$ severities ${0, 0.1, \ldots, 0.9, 0.95, 0.97, 0.99}$, plus a fault-free cell. The \textbf{omnidirectional sweep} extends FT-CPG's evaluation schedule~\cite{zhang2025ftcpg} over severities ${0, 0.3, 0.5, 0.7, 0.8, 0.9, 0.95, 0.99}$. After a \SI{4}{s} forward lead-in, its \SI{25}{s} episodes use \SI{5}{s} segments of $v_x$, $v_y=\SI{0.4}{m/s}$ and $\omega_z=\SI{0.6}{rad/s}$. A fourth segment combines $v_x$ and $\omega_z$ with one random sign per episode, covering backward and sideways walking and turning in place.
Both sweeps command $v_x=\SI{0.5}{m/s}$ for every arm and jitter the start pose and fault onset per episode, giving independent trials. The straight-line sweep verifies at every step that the policy observed the command set, whereas the omnidirectional sweep leaves the simulator's sampler running, which swaps in a random command for one step at \SI{10}{s}. The command-verified \textbf{robustness test} adds one change of friction, payload, PD gains, $\dot q$ noise or a sideways push to a weakened calf.

%% file: sections/05_experiments_and_results.tex
\section{Experiments and Results} \label{sec:results}

\subsection{Simulation: the trot survives the fault}
\label{sec:results:sim}

\begin{figure*}[t]
\centering
\includegraphics[width=0.95\textwidth]{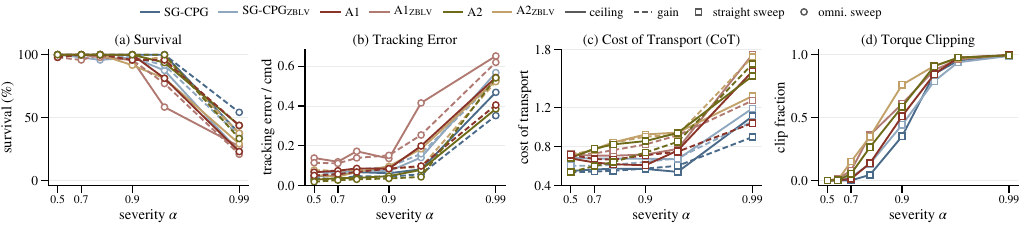}
\caption{Performance of different policies on the evaluation protocol. Hollow circles are the omnidirectional schedule of Sec.~\ref{sec:protocol}, hollow squares the straight-line sweep at $v_x = \SI{0.5}{m/s}$.
(a) Survival (\%) over $36$ episodes (b) Tracking error averaged over the four commanded segments of the  omnidirectional schedule (c) Cost of transport (CoT) (d) clip fraction, for \texttt{ceiling} arms only.
Ablations A1 (no severity observation) and A2 (no amplitude gate) appear on both bases.}
\label{fig:severity}
\vspace{-4mm}
\end{figure*}

\textbf{Gait.}
On the straight-line sweep trot wins the majority vote at every severity for all four policies, and at $\sev = 0.99$ at least $8$ of $12$ cells per policy still trot.
Cost of transport stays within $13\%$ of its $\sev = 0.5$ value through $0.95$ before rising $1.7$--$2.0\times$ at $0.99$ (Fig.~\ref{fig:severity}c).

\textbf{Survival with tracking.}
On the straight-line sweep no policy loses more than two of $36$ episodes per severity up to $\sev = 0.95$, whereas at $0.99$ the four keep $20$--$28$.
On the omnidirectional sweep at the same forward command, survival at $\sev = 0.95$ is $100\%$ for the flown pair and $87.5$--$91.7\%$ for the ZBLV pair, and at $0.99$ it drops to $37.5$--$54.2\%$ and $22.9$--$35.4\%$, so straight-line tests overstate what a weakened robot can do (Fig.~\ref{fig:severity}a).
Upright episodes still follow the command closely: tracking error stays within $9.3\%$ of command through $\sev = 0.9$ and $16.0\%$ through $0.95$, before reaching $35$--$57\%$ at $0.99$ (Fig.~\ref{fig:severity}b).
Sideways stepping is the weakest axis and turning the strongest: over that range the flown \texttt{ceiling} arm holds $7.2$--$14.3\%$ and at worst $4.3\%$ error.
Tracking is measured only on episodes completing a segment, so it hides falls and is read beside survival.

\subsection{The two mechanisms are physically distinct}
\label{sec:results:mech}

If the dead band of Sec.~\ref{sec:method:severity} is real, a lowered ceiling should saturate the faulted joint only once demand reaches the cap, whereas scaled gains should rarely saturate it.
Under \texttt{ceiling} the lowered cap leaves the tested joint's median clip fraction at zero through $\sev = 0.5$ and clips $35.1\%$ of its steps at $0.9$, whereas both \texttt{gain} policies hold zero throughout (Fig.~\ref{fig:severity}d).

\subsection{Robustness beyond the fault}
\label{sec:results:robust}

\input{tables/t4_robust}

A weakened robot also meets slippery ground, extra load, servo and sensor errors and shoves, so we walk the four policies at \SI{0.5}{m/s} under one such change at a time (Table~\ref{tab:robust}).
Lower friction, stiffer gains and joint-velocity noise cost no faulted episode of $40$, though noise alone more than doubles the pooled $v_x$ error to \SI{0.138}{m/s}.
Added load and softened gains are the exceptions.
\texttt{ceiling} without base velocity falls in every front-left episode from $+5$~kg, the top of its trained mass range, whereas the flown \texttt{ceiling} policy gives way on its rear-right calf instead.
The robot's fault sits on the front-right calf, which this grid does not test, so neither calf-specific failure should be read as a limit of the hardware runs.

\subsection{What can be estimated?}
\label{sec:results:est}

\textbf{The estimate can drive the controller.}
Substituting the analytic estimator's severity for the injected one on a three-axis schedule costs the flown \texttt{ceiling} arm nothing until its hardest rear-right condition, where it keeps $80\%$, and the \texttt{gain} arm only its front-left level, at $70\%$.

\begin{table}[t]\centering\footnotesize
\renewcommand{\arraystretch}{1.15}
\caption{Severity MAE on held-out windows under training commands (Sec.~\ref{sec:method:est}).
Learned models are trained once each. Bold marks the lowest faulted MAE per mechanism.}
\label{tab:est}
\setlength{\tabcolsep}{3pt}
\begin{tabular}{lrrcrrr}\hline
 & \multicolumn{2}{c}{\texttt{gain}} & & \multicolumn{2}{c}{\texttt{ceiling}} & \\ \cline{2-3}\cline{5-6}
estimator & faulted & healthy & & faulted & healthy & params \\ \hline
predict $0$ & 0.633 & 0.000 & & 0.720 & 0.000 & 0 \\
analytic & 0.182 & 0.627 & & \textbf{0.076} & 0.620 & 0 \\
Transformer-10 & \textbf{0.152} & 0.009 & & 0.310 & 0.021 & 246k \\
CNN-30 & 0.170 & 0.009 & & 0.330 & 0.020 & 55k \\
MLP-30 & 0.214 & 0.010 & & 0.456 & 0.023 & 74k \\
MLP-1 & 0.232 & 0.010 & & 0.492 & 0.022 & 38k \\ \hline
\end{tabular}
\vspace{-4mm}
\end{table}

\textbf{The mechanism decides which estimator is right.}
Under \texttt{ceiling} the unclipped analytic estimator beats every learned model on faulted joints at least fourfold (Table~\ref{tab:est}), with an MAE of $0.007$ on joint windows reaching $95\%$ of their current cap.
Under \texttt{gain}, where only $1.3\%$ of joint windows reach that level against $6.0\%$ under \texttt{ceiling}, the two temporal models edge ahead of it.
On healthy joints, however, the analytic estimator fails by reading any lightly loaded joint as weak.

\textbf{Temporal structure, not history length, buys accuracy.}
Feeding a dense layer thirty frames instead of one lowers faulted MAE by $7\%$, whereas convolving them or attending over the last ten lowers it by a further $21$--$32\%$.

\textbf{On the robot the estimator substitutes for a hand-set severity, while misreading it.}
Feeding the analytic estimator's output to \sgcpg{} in place of a hand-set $\sev = 0.85$ on the front-right calf, both runs walk without falling at speeds matching two hand-set runs of the same cell.
Yet the estimate averages $0.363$ and $0.385$ against the true $0.85$ and reads $0.10$ on the eleven healthy joints, so the policy walked on a severity far milder than the one applied.
The estimate is therefore usable in the loop without being accurate, though at $\gd = 0.25$ the residual has little authority to misapply.

\subsection{Ablations} \label{sec:ablations}

Besides opening the residual gate $\gd$, severity enters $\pid$ as an observation and shortens a weakened leg's stride through the amplitude gate.
We isolate each path with its own retrained arm per mechanism \emph{and per healthy base}, so every ablation is read against a control differing from it in one path alone.
\textbf{A1}, the severity-blind residual, removes $\sev$ from $\pid$'s observation, while $\gd$ and the amplitude gate of~\eqref{eq:gate} still act on the injected severity.
\textbf{A2}, the no-amplitude-gate arm, keeps that observation but leaves $\mu_i$ unscaled, so nothing shortens a weakened leg's stride to ease its joint's torque.

\textbf{Either path removed, the weak joint is asked for more.}
Under \texttt{ceiling} at $\sev = 0.8$ the weakened joint's median clip fraction rises from $4.4\%$ for \sgcpg{} to $13.6\%$ for A1 and $26.3\%$ for A2, and on the ZBLV base from $13.3\%$ to $36.0\%$ and $34.8\%$ (Fig.~\ref{fig:severity}d).
Both paths exist to lower what the controller demands of a weakened actuator, and removing either raises that demand on both bases and at every severity.
Energy follows the same pattern, and not only under \texttt{ceiling}: at $\sev = 0.95$ A1 and A2 both raise cost of transport above their own control under each mechanism and on each base (Fig.~\ref{fig:severity}c).

\textbf{Only the severity observation costs survival.}
At $\sev = 0.95$ A1 keeps $81.2$ and $95.8\%$ of its omnidirectional episodes against $100\%$ for \sgcpg{} under \texttt{ceiling} and \texttt{gain}, and $58.3$ and $77.1\%$ against $87.5$ and $91.7\%$ on the ZBLV base (Fig.~\ref{fig:severity}a).
It also tracks worst in all four cells, at $19.9$ and $9.8\%$ mean error against $8.1$ and $5.7\%$, and $41.6$ and $25.4\%$ against $14.0$ and $16.0\%$ (Fig.~\ref{fig:severity}b), whereas A2 costs at most one episode in sixteen on either base.
Removing the gate therefore trades joint load for tracking rather than degrading the controller, whereas removing the severity observation degrades both.

%% file: tables/t4_robust.tex
\definecolor{robustfail}{RGB}{176,122,114}
\begin{table}[t]\centering\footnotesize
\renewcommand{\arraystretch}{1.15}
\caption{Surviving episodes of $40$ per weakened calf under one disturbance at a time, shaded by the shortfall from $40$, with the last column averaging the $v_x$ error of survivors.}
\label{tab:robust}
\setlength{\tabcolsep}{3.4pt}
\scriptsize
\begin{tabular}{l*{8}{>{\centering\arraybackslash}p{13pt}}>{\centering\arraybackslash}p{22pt}}\hline
 & \multicolumn{4}{c}{\sgcpg{}} & \multicolumn{4}{c}{\sgcpg{}$_{\mathrm{ZBLV}}$} & \\ \cline{2-5}\cline{6-9}
 & \multicolumn{2}{c}{\texttt{ceiling}} & \multicolumn{2}{c}{\texttt{gain}} & \multicolumn{2}{c}{\texttt{ceiling}} & \multicolumn{2}{c}{\texttt{gain}} & Mean \\
condition & FL & RR & FL & RR & FL & RR & FL & RR & RMSE \\ \hline
nominal & 40 & 40 & 40 & 40 & 39 & 40 & 40 & 40 & 0.057 \\ \hline
friction $0.35$ & 40 & 40 & 40 & 40 & 40 & 40 & 40 & 40 & 0.061 \\ \hline
payload \SI{+5}{kg} & 40 & \cellcolor{robustfail!21}25 & 40 & 40 & \cellcolor{robustfail!55}0 & 40 & 40 & 40 & 0.051 \\
payload \SI{+7}{kg} & 40 & \cellcolor{robustfail!32}17 & \cellcolor{robustfail!10}33 & 40 & \cellcolor{robustfail!55}0 & 40 & 40 & 40 & 0.058 \\ \hline
gains $\times 0.65$ & 40 & \cellcolor{robustfail!44}8 & \cellcolor{robustfail!14}30 & 40 & \cellcolor{robustfail!55}0 & 40 & 40 & 40 & 0.061 \\
gains $\times 1.35$ & 40 & 40 & 40 & 40 & 40 & 40 & 40 & 40 & 0.066 \\ \hline
$\dot q$ noise & 40 & 40 & 40 & 40 & 40 & 40 & 40 & 40 & 0.138 \\ \hline
push \SI{1.0}{m/s} & \cellcolor{robustfail!6}36 & 39 & 40 & 40 & \cellcolor{robustfail!21}25 & 40 & \cellcolor{robustfail!10}33 & 40 & 0.073 \\
push \SI{1.5}{m/s} & \cellcolor{robustfail!18}27 & \cellcolor{robustfail!16}28 & \cellcolor{robustfail!28}20 & \cellcolor{robustfail!25}22 & \cellcolor{robustfail!45}7 & \cellcolor{robustfail!16}28 & \cellcolor{robustfail!40}11 & \cellcolor{robustfail!40}11 & 0.080 \\ \hline
\end{tabular}
\end{table}

%% file: sections/06_hardware.tex
\section{Hardware Deployment}
\label{sec:results:hw}

We run \texttt{ceiling} \sgcpg{} on a Unitree Go2 with the front-right calf at a hand-set $\sev$, emulating each cap by clamping position errors, as the motor command has no torque-limit field.
At $\sev = 0.85$ the motor's torque estimate exceeds the $\SI{6.81}{N\,m}$ calf cap on $23$--$44\%$ of ticks, so hardware severities are milder than simulated ones.

\begin{figure*}[t]
\centering
\includegraphics[width=0.95\textwidth]{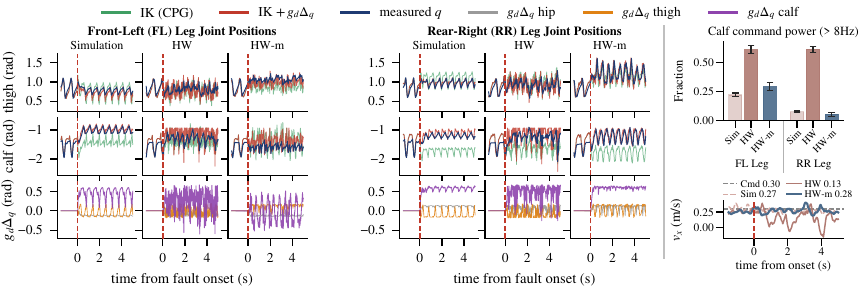}
\caption{Healthy-leg adaptation to a weakened FR calf at $\sev = 0.85$ under \texttt{ceiling} mechanism.
(Left and Center): Joint positions of FL and RR legs in simulation, on hardware (HW) and after a plant-matched retrain (HW-m), each giving measured $q$, IK target (with and without) applied $\gd\dres_{q}$.
(Right): Calf-command power above \SI{8}{Hz} and robot forward speed.}
\label{fig:hwadapt}
\vspace{-5mm}
\end{figure*}

\textbf{Healthy transfer.}
Without a fault \sgcpg{} reduces to $\pih$, which reaches $87$--$100\%$ of commanded speed by tape up to \SI{0.7}{m/s}, after which speed rises little while the thigh spends longer at its torque clamp.
Leg odometry agrees with tape to $0.96$--$1.04\times$ here, but under fault it is our only forward-speed measure and, as the policy's velocity input, not independent.

\begin{table}[t]\centering\footnotesize
\renewcommand{\arraystretch}{1.15}
\caption{SG-CPG hardware envelope showcasing number of falls for ceiling fault with severity $\sev$ on FL calf.}
\label{tab:envelope}
\setlength{\tabcolsep}{4pt}
\begin{tabular}{lll}\hline
$\sev$ & forward & yaw \\ \hline
0.85--0.87 & 0/13 (to $v_x$ 1.0) & 0/4 (to $|\omega_z|$ 0.6) \\
0.88--0.90 & 0/3 & 0/3 (incl.\ $|\omega_z|$ 1.0) \\
0.91--0.93 & 1/4 & 0/2 (incl.\ $|\omega_z|$ 0.8) \\
0.94 & 1/8 (falls at $v_x$ 0.8) & 1/1 (falls at $|\omega_z|$ 0.6) \\
0.95--0.97 & 2/4 & 5/5 (all at $\sev$ 0.95) \\ \hline
\end{tabular}
\vspace{-4mm}
\end{table}

\textbf{The envelope reaches $\sev = 0.93$.}
Across $\sev = 0.85$--$0.93$, \sgcpg{} falls in one of $20$ forward runs to $\SI{1.0}{m/s}$ and none of $9$ turning runs to $\SI{1.0}{rad/s}$ (Table~\ref{tab:envelope}).
Above $\sev = 0.93$ frequent falls leave only survival to report.

\textbf{On a weakened thigh the residual decides between walking and falling.}
We also weaken the front-right \emph{thigh} to $\sev = 0.80$ and run the plant-matched policy at $v_x = \SI{0.3}{m/s}$, at full residual authority and with the residual gated off.
With it, four of four runs stay upright at a mean \SI{0.26}{m/s} by odometry, tilting $2.6$--$3.2^\circ$; gated off, both tip past the $34.4^\circ$ cut-out, $2.5$ and \SI{4.0}{s} after onset.

\textbf{The amplitude gate asks less of the weak leg.}
At $\sev = 0.85$ and $v_x = \SI{0.3}{m/s}$ even $\pih$ alone stands, so we compare two runs of it, three with the gate added and six of \sgcpg{}, none falling.
The amplitude gate alone halves mean body tilt from $6.8$--$7.2^\circ$ to $3.6$--$3.9^\circ$ and the share of ticks with the calf's commanded torque clamped at the cap from $76$--$77\%$ to $37$--$40\%$.
This matches the relief under a simulated ceiling but costs speed.
The down-scaled residual recovers most of that speed and trims tilt to $2.4$--$3.1^\circ$, while the commanded calf torque stays clamped on $34$--$44\%$ of ticks.

\textbf{Matching the actuator model makes the adaptation realisable.}
Identifying the actuators from a chirp adds the joint friction the training model lacks and re-centres stiffness.
The residual answers a weakened calf by extending the healthy legs' calves and holding them there, $0.56$--\SI{0.58}{rad} of $\gd\dres_{q}$ across the stride in simulation (Fig.~\ref{fig:hwadapt}).
On the robot the same policy's command chatters: the rear-right leg's share of calf-command power above \SI{8}{Hz} rises from $0.08$ to $0.61$ and forward speed falls from $0.27$ to \SI{0.13}{m/s}.
Retrained on the matched model it returns to $0.05$, within the $0.04$ of its own simulation, and speed holds at \SI{0.28}{m/s}.


%% file: sections/07_conclusion_and_futurework.tex
\section{Conclusion} \label{sec:conclusion}
We presented SG-CPG, a single robot-level controller for quadruped locomotion under \emph{continuous} actuator degradation. SG-CPG builds on a frozen healthy CPG policy with a severity-gated residual that enables whole-body re-coordination and a severity-gated oscillator amplitude that reduces the weakened leg's motion demand. The gait therefore adapts progressively as actuator capability decreases, without requiring a separate policy for each fault severity or affected limb. Across simulation and real-robot experiments, SG-CPG demonstrates that gradual torque loss can be accommodated while retaining the nominal gait structure. In simulation, it preserves trotting under a $95\%$ reduction in torque ceiling with approximately $8\%$ velocity-tracking error while maintaining omnidirectional motion. On a Unitree Go2, SG-CPG sustains locomotion under up to $93\%$ emulated torque-ceiling reduction. Beyond this range, the effects of degradation become increasingly apparent, with the robot yawing and rolling toward the degraded limb and falling more frequently. Overall, these results show that continuous actuator degradation can be handled through gradual, severity-dependent modification of a nominal gait rather than a discrete switch to a fault- or limb-specific locomotion policy. SG-CPG provides a framework for extending fault-tolerant locomotion beyond catastrophic failures toward progressive actuator degradation expected during long-term robot operation.

%% file: root.bib
@article{chang2009wholelimb,
  author  = {Chang, Young-Hui and Auyang, Arick G. and Scholz, John P. and Nichols, T. Richard},
  title   = {Whole limb kinematics are preferentially conserved over individual joint kinematics after peripheral nerve injury},
  journal = {Journal of Experimental Biology},
  volume  = {212},
  number  = {21},
  pages   = {3511--3521},
  year    = {2009},
  doi     = {10.1242/jeb.033886}
}

@article{chang2018progressive,
  author  = {Chang, Young-Hui and Housley, Stephen N. and Hart, Kerry S. and Nardelli, Paul and Nichols, Richard T. and Maas, Huub and Cope, Timothy C.},
  title   = {Progressive adaptation of whole-limb kinematics after peripheral nerve injury},
  journal = {Biology Open},
  volume  = {7},
  number  = {8},
  pages   = {bio028852},
  year    = {2018},
  doi     = {10.1242/bio.028852}
}

@article{farid2018fractional,
  author  = {Farid, Yousef and Majd, Vahid Johari and Ehsani-Seresht, Abbas},
  title   = {Fractional-order active fault-tolerant force-position controller design for the legged robots using saturated actuator with unknown bias and gain degradation},
  journal = {Mechanical Systems and Signal Processing},
  volume  = {104},
  pages   = {465--486},
  year    = {2018},
  doi     = {10.1016/j.ymssp.2017.11.010}
}

@article{cruz2017gait,
  author  = {Cruz-Jimenez, Maricarmen},
  title   = {Normal Changes in Gait and Mobility Problems in the Elderly},
  journal = {Physical Medicine and Rehabilitation Clinics of North America},
  volume  = {28},
  number  = {4},
  pages   = {713--725},
  year    = {2017},
  doi     = {10.1016/j.pmr.2017.06.005}
}

@article{ijspeert2008review,
  author  = {Ijspeert, Auke Jan},
  title   = {Central pattern generators for locomotion control in animals and robots: A review},
  journal = {Neural Networks},
  volume  = {21}, number = {4}, pages = {642--653}, year = {2008}
}

@article{bellegarda2022cpgrl,
  author  = {Bellegarda, Guillaume and Ijspeert, Auke},
  title   = {{CPG-RL}: Learning central pattern generators for quadruped locomotion},
  journal = {IEEE Robotics and Automation Letters},
  volume  = {7}, number = {4}, pages = {12547--12554}, year = {2022}
}

@inproceedings{bellegarda2024visual,
  author    = {Bellegarda, Guillaume and Shafiee, Milad and Ijspeert, Auke},
  title     = {Visual {CPG-RL}: Learning central pattern generators for visually-guided quadruped locomotion},
  booktitle = {Proc. IEEE Int. Conf. Robotics and Automation (ICRA)},
  pages     = {1420--1427}, year = {2024},
  doi       = {10.1109/ICRA57147.2024.10611128}
}

@inproceedings{bellegarda2024allgaits,
  author    = {Bellegarda, Guillaume and Shafiee, Milad and Ijspeert, Auke},
  title     = {{AllGaits}: Learning all quadruped gaits and transitions},
  booktitle = {Proc. IEEE Int. Conf. Robotics and Automation (ICRA)},
  pages     = {15929--15935}, year = {2025},
  doi       = {10.1109/ICRA55743.2025.11127285}
}

@article{shafiee2024viability,
  author  = {Shafiee, Milad and Bellegarda, Guillaume and Ijspeert, Auke},
  title   = {Viability leads to the emergence of gait transitions in learning agile quadrupedal locomotion on challenging terrains},
  journal = {Nature Communications},
  volume  = {15}, pages = {3073}, year = {2024}
}

@article{luo2023ftnet,
  author  = {Luo, Zeren and Xiao, Erdong and Lu, Peng},
  title   = {{FT-Net}: Learning failure recovery and fault-tolerant locomotion for quadruped robots},
  journal = {IEEE Robotics and Automation Letters},
  volume  = {8}, number = {12}, pages = {8414--8421}, year = {2023}
}

@article{zhang2025ftcpg,
  author  = {Zhang, Pei and Hua, Zhaobo and Qiu, Qiyu and Ding, Jinliang},
  title   = {{FT-CPG}: Learning central pattern generators for fault-tolerant quadruped locomotion under multi-joint failures},
  journal = {IEEE Robotics and Automation Letters},
  volume  = {10}, number = {7}, pages = {6936--6943}, year = {2025},
  doi = {10.1109/LRA.2025.3572772}
}

@misc{liu2022limping,
  author        = {Liu, Dikai and Zhang, Tianwei and Yin, Jianxiong and See, Simon},
  title         = {Saving the limping: Fault-tolerant quadruped locomotion via reinforcement learning},
  year          = {2022},
  eprint        = {2210.00474}, archivePrefix = {arXiv}, primaryClass = {cs.RO},
  note          = {arXiv:2210.00474}
}

@inproceedings{kim2024masking,
  author    = {Kim, Mincheol and Shin, Ukcheol and Kim, Jung-Yup},
  title     = {Learning quadrupedal locomotion with impaired joints using random joint masking},
  booktitle = {Proc. IEEE Int. Conf. Robotics and Automation (ICRA)},
  pages     = {9751--9757}, year = {2024}
}

@inproceedings{lee2025dreamflex,
  author    = {Lee, Seunghyun and Nahrendra, I Made Aswin and Lee, Dongkyu and Yu, Byeongho and Oh, Minho and Myung, Hyun},
  title     = {{DreamFLEX}: Learning fault-aware quadrupedal locomotion controller for anomaly situation in rough terrains},
  booktitle = {Proc. IEEE Int. Conf. Robotics and Automation (ICRA)},
  pages     = {16001--16007}, year = {2025},
  doi       = {10.1109/ICRA55743.2025.11127805}
}

@misc{xu2025acl,
  author        = {Xu, Tianyu and Cheng, Yaoyu and Shen, Pinxi and Zhao, Lin},
  title         = {{AcL}: Action learner for fault-tolerant quadruped locomotion control},
  year          = {2025},
  eprint        = {2503.21401}, archivePrefix = {arXiv}, primaryClass = {cs.RO},
  note          = {arXiv:2503.21401}
}

@misc{gravina2026gaittiming,
  author        = {Gravina, Giovanbattista and Rossini, Luca and Rizzardo, Carlo
                   and Laurenzi, Arturo and Tsagarakis, Nikos},
  title         = {Learning Fault-Tolerant Locomotion with Adaptive Gait Timing},
  year          = {2026},
  eprint        = {2608.07328},
  archivePrefix = {arXiv},
  primaryClass  = {cs.RO},
  note          = {Accepted at IEEE/RSJ International Conference on Intelligent Robots and Systems (IROS), 2026}
}

@inproceedings{hou2024multitask,
  author    = {Hou, Taixian and Tu, Jiaxin and Gao, Xiaofei and Dong, Zhiyan and Zhai, Peng and Zhang, Lihua},
  title     = {Multi-task learning of active fault-tolerant controller for leg failures in quadruped robots},
  booktitle = {Proc. IEEE Int. Conf. Robotics and Automation (ICRA)},
  pages     = {9758--9764}, year = {2024},
  doi       = {10.1109/ICRA57147.2024.10610151}
}

@article{bongard2006resilient,
  author  = {Bongard, Josh and Zykov, Victor and Lipson, Hod},
  title   = {Resilient machines through continuous self-modeling},
  journal = {Science},
  volume  = {314}, number = {5802}, pages = {1118--1121}, year = {2006}
}

@article{cully2015adapt,
  author  = {Cully, Antoine and Clune, Jeff and Tarapore, Danesh and Mouret, Jean-Baptiste},
  title   = {Robots that can adapt like animals},
  journal = {Nature},
  volume  = {521}, pages = {503--507}, year = {2015}
}

@article{koos2013tresilience,
  author  = {Koos, Sylvain and Cully, Antoine and Mouret, Jean-Baptiste},
  title   = {Fast damage recovery in robotics with the {T-Resilience} algorithm},
  journal = {The International Journal of Robotics Research},
  volume  = {32}, number = {14}, pages = {1700--1723}, year = {2013}
}

@inproceedings{wu2023adapt,
  author    = {Wu, Xinyuan and Dong, Wentao and Lai, Hang and Yu, Yong and Wen, Ying},
  title     = {Adaptive control strategy for quadruped robots in actuator degradation scenarios},
  booktitle = {Proc. 5th Int. Conf. Distributed Artificial Intelligence (DAI)},
  pages     = {1--13}, publisher = {ACM}, year = {2023},
  doi       = {10.1145/3627676.3627686}
}

@article{lee2020challenging,
  author  = {Lee, Joonho and Hwangbo, Jemin and Wellhausen, Lorenz and Koltun, Vladlen and Hutter, Marco},
  title   = {Learning quadrupedal locomotion over challenging terrain},
  journal = {Science Robotics},
  volume  = {5}, number = {47}, pages = {eabc5986}, year = {2020}
}

@inproceedings{kumar2021rma,
  author    = {Kumar, Ashish and Fu, Zipeng and Pathak, Deepak and Malik, Jitendra},
  title     = {{RMA}: Rapid motor adaptation for legged robots},
  booktitle = {Proc. Robotics: Science and Systems (RSS)},
  year      = {2021}
}

@misc{silver2018residual,
  author        = {Silver, Tom and Allen, Kelsey and Tenenbaum, Josh and Kaelbling, Leslie},
  title         = {Residual policy learning},
  year          = {2018},
  eprint        = {1812.06298}, archivePrefix = {arXiv}, primaryClass = {cs.RO},
  note          = {arXiv:1812.06298}
}

@misc{wan2026thermal,
  author        = {Wan, Yuhang and others},
  title         = {Learning to balance motor thermal safety and quadrupedal locomotion performance with residual policy},
  year          = {2026},
  eprint        = {2605.27046}, archivePrefix = {arXiv}, primaryClass = {cs.RO},
  note          = {arXiv:2605.27046}
}

@article{bouyer2001plasticity,
  author  = {Bouyer, Laurent J. G. and Whelan, Patrick J. and Pearson, Keir G. and Rossignol, Serge},
  title   = {Adaptive locomotor plasticity in chronic spinal cats after ankle extensors neurectomy},
  journal = {Journal of Neuroscience},
  volume  = {21}, number = {10}, pages = {3531--3541}, year = {2001}
}

@article{bellicoso2018realworld,
  author  = {Bellicoso, C. Dario and others},
  title   = {Advances in real-world applications for legged robots},
  journal = {Journal of Field Robotics},
  volume  = {35}, number = {8}, pages = {1311--1326}, year = {2018},
  doi     = {10.1002/rob.21839}
}

@inproceedings{gehring2021anymalhvdc,
  author    = {Gehring, Christian and others},
  title     = {{ANYmal} in the field: Solving industrial inspection of an offshore {HVDC} platform with a quadrupedal robot},
  booktitle = {Field and Service Robotics},
  series    = {Springer Proceedings in Advanced Robotics}, volume = {16},
  publisher = {Springer}, pages = {247--260}, year = {2021},
  doi       = {10.1007/978-981-15-9460-1_18}
}

@article{wensing2017proprioceptive,
  author  = {Wensing, Patrick M. and Wang, Albert and Seok, Sangok and Otten, David and Lang, Jeffrey and Kim, Sangbae},
  title   = {Proprioceptive actuator design in the {MIT} {Cheetah}: Impact mitigation and high-bandwidth physical interaction for dynamic legged robots},
  journal = {IEEE Transactions on Robotics},
  volume  = {33}, number = {3}, pages = {509--522}, year = {2017},
  doi     = {10.1109/TRO.2016.2640183}
}

@article{shin2025motorregion,
  author  = {Shin, Young-Ha and Song, Tae-Gyu and Ji, Gwanghyeon and Park, Hae-Won},
  title   = {Reinforcement learning for high-speed quadrupedal locomotion with motor operating region constraints: Mitigating motor model discrepancies through torque clipping in realistic motor operating region},
  journal = {IEEE Robotics \& Automation Magazine},
  volume  = {32}, number = {2}, pages = {49--59}, year = {2025},
  doi     = {10.1109/MRA.2024.3487322}
}

@misc{qian2026thermalaware,
  author        = {Qian, Letian and Wan, Yuhang and Wang, Shuhan and Luo, Xin},
  title         = {Learning thermal-aware locomotion policies for an electrically-actuated quadruped robot},
  year          = {2026},
  eprint        = {2603.01631}, archivePrefix = {arXiv}, primaryClass = {cs.RO},
  note          = {arXiv:2603.01631}
}

@article{ijspeert2014biorobotics,
  author  = {Ijspeert, Auke J.},
  title   = {Biorobotics: Using robots to emulate and investigate agile locomotion},
  journal = {Science},
  volume  = {346},
  number  = {6206},
  pages   = {196--203},
  year    = {2014},
  doi     = {10.1126/science.1254486}
}

@article{hughes1957coordination,
  author  = {Hughes, G. M.},
  title   = {The co-ordination of insect movements. {II}. {The} effect of limb amputation and the cutting of commissures in the cockroach (\emph{{Blatta orientalis}})},
  journal = {Journal of Experimental Biology},
  volume  = {34}, number = {3}, pages = {306--333}, year = {1957},
  doi     = {10.1242/jeb.34.3.306}
}

@article{kane2025spiders,
  author  = {Kane, Suzanne Amador and Quinn, Brooke L. and Wu, Xuanyi Kris and Xi, Sarah Y. and Ochs, Michael F. and Hsieh, S. Tonia},
  title   = {Unsupervised learning reveals rapid gait adaptation after leg loss and regrowth in spiders},
  journal = {Journal of Experimental Biology},
  volume  = {228}, number = {12}, pages = {jeb250243}, year = {2025},
  doi     = {10.1242/jeb.250243}
}

@article{goldner2018lameness,
  author  = {Goldner, Birte and Fischer, Stefanie and Nolte, Ingo and Schilling, Nadja},
  title   = {Kinematic adaptions to induced short-term pelvic limb lameness in trotting dogs},
  journal = {BMC Veterinary Research},
  volume  = {14}, number = {1}, pages = {183}, year = {2018},
  doi     = {10.1186/s12917-018-1484-2}
}

@article{pearson1999denervation,
  author  = {Pearson, K. G. and Fouad, K. and Misiaszek, J. E.},
  title   = {Adaptive changes in motor activity associated with functional recovery following muscle denervation in walking cats},
  journal = {Journal of Neurophysiology},
  volume  = {82}, number = {1}, pages = {370--381}, year = {1999},
  doi     = {10.1152/jn.1999.82.1.370}
}

@article{brown1911intrinsic,
  author  = {Brown, T. Graham},
  title   = {The intrinsic factors in the act of progression in the mammal},
  journal = {Proceedings of the Royal Society of London. Series B, Containing Papers of a Biological Character},
  volume  = {84},
  number  = {572},
  pages   = {308--319},
  year    = {1911},
  doi     = {10.1098/rspb.1911.0077}
}

@article{kiehn2016decoding,
  author  = {Kiehn, Ole},
  title   = {Decoding the organization of spinal circuits that control locomotion},
  journal = {Nature Reviews Neuroscience},
  volume  = {17},
  number  = {4},
  pages   = {224--238},
  year    = {2016},
  doi     = {10.1038/nrn.2016.9}
}

@article{rossignol2006dynamic,
  author  = {Rossignol, Serge and Dubuc, R{\'e}jean and Gossard, Jean-Pierre},
  title   = {Dynamic sensorimotor interactions in locomotion},
  journal = {Physiological Reviews},
  volume  = {86},
  number  = {1},
  pages   = {89--154},
  year    = {2006},
  doi     = {10.1152/physrev.00028.2005}
}

@article{grillner2020current,
  author  = {Grillner, Sten and El Manira, Abdeljabbar},
  title   = {Current principles of motor control, with special reference to vertebrate locomotion},
  journal = {Physiological Reviews},
  volume  = {100},
  number  = {1},
  pages   = {271--320},
  year    = {2020},
  doi     = {10.1152/physrev.00015.2019}
}

@article{carrier1997neurectomy,
  author  = {Carrier, L. and Brustein, E. and Rossignol, S.},
  title   = {Locomotion of the hindlimbs after neurectomy of ankle flexors in intact and spinal cats: Model for the study of locomotor plasticity},
  journal = {Journal of Neurophysiology},
  volume  = {77}, number = {4}, pages = {1979--1993}, year = {1997},
  doi     = {10.1152/jn.1997.77.4.1979}
}

@article{ren2015chaotic,
  author  = {Ren, Guanjiao and Chen, Weihai and Dasgupta, Sakyasingha and Kolodziejski, Christoph and W{\"o}rg{\"o}tter, Florentin and Manoonpong, Poramate},
  title   = {Multiple chaotic central pattern generators with learning for legged locomotion and malfunction compensation},
  journal = {Information Sciences},
  volume  = {294},
  pages   = {666--682},
  year    = {2015},
  doi     = {10.1016/j.ins.2014.05.001}
}
